%% file: main.tex
\documentclass[conference]{IEEEtran}
\usepackage{blindtext, graphicx}

\ifCLASSINFOpdf
\else
\fi

\usepackage[utf8]{inputenc}
\usepackage{multirow}

\usepackage{ifpdf}
\usepackage{epstopdf}
\DeclareGraphicsExtensions{.eps}
\usepackage[font=footnotesize]{subfig}

\usepackage{comment}
\usepackage{amsmath,amsthm,amsfonts,amssymb}
\usepackage[noadjust]{cite}
\usepackage{algorithm, algorithmic}
\usepackage{graphics}
\usepackage{color}

\usepackage{caption}

\usepackage[colorinlistoftodos, textwidth=4cm]{todonotes}

\IEEEoverridecommandlockouts

\begin{document}
\title{Probabilistic Load Forecasting via Point Forecast Feature Integration} 

\author{	
\IEEEauthorblockN{
Qicheng~Chang$^{1}$,
Yishen~Wang$^{1}$,
Xiao~Lu$^{2}$,
Di~Shi$^{1}$,
Haifeng~Li$^{2}$,
Jiajun~Duan$^{1}$,
Zhiwei~Wang$^{1}$
}
\IEEEauthorblockA{$^{1}$GEIRI North America, San Jose, CA, USA}
\IEEEauthorblockA{$^{2}$State Grid Jiangsu Electric Power Company Ltd., Nanjing, Jiangsu, China}
\IEEEauthorblockA{Email: yishen.wang@geirina.net}
\vspace{-30pt}
\thanks{This work is funded by SGCC Science and Technology Program under contract no. SGSDYT00FCJS1700676.
}}

\maketitle

\begin{abstract}
Short-term load forecasting is a critical element of power systems energy management systems. In recent years, probabilistic load forecasting (PLF) has gained increased attention for its ability to provide uncertainty information that helps to improve the reliability and economics of system operation performances. This paper proposes a two-stage probabilistic load forecasting framework by integrating point forecast as a key probabilistic forecasting feature into PLF. In the first stage, all related features are utilized to train a point forecast model and also obtain the feature importance. In the second stage the forecasting model is trained, taking into consideration point forecast features, as well as selected feature subsets. During the testing period of the forecast model, the final probabilistic load forecast results are leveraged to obtain both point forecasting and probabilistic forecasting. Numerical results obtained from ISO New England demand data demonstrate the effectiveness of the proposed approach in the hour-ahead load forecasting, which uses the gradient boosting regression for the point forecasting and quantile regression neural networks for the probabilistic forecasting.
\end{abstract}


\IEEEpeerreviewmaketitle

\section{Introduction}

Short-term load forecasting (STLF) aims to provide accurate future load setpoints for economic and reliable system operations. STLF has been a standard for most practical energy management (EMS) applications and an active research area for decades \cite{amjady_2001_forecast, rahman_1988_forecast, hippert_2001_nn, kang_2004_review}. Traditionally, STLF is mainly conducted with point forecasting, which outputs a deterministic estimation to represent the expected load for the targeted time. Time series analysis \cite{amjady_2001_forecast}, expert systems \cite{rahman_1988_forecast}, artificial neural networks \cite{hippert_2001_nn} and multiple linear regression \cite{kang_2004_review} have all achieved satisfactory results in the past. 

Recent advancements in the field of artificial intelligence have resulted in new machine learning applications for energy forecasting. The Global Energy Forecasting Competition 2012 (GEFCOM2012) \cite{hong_2014_gefcom2012} was devoted to state-of-the-art point forecasting techniques for wind and load, as well making available to the public benchmark datasets that would be of specific interest to industry practitioners and academic researchers. In this competition, a number of techniques, including data cleansing, hierarchical forecasting, special days forecasting, temperature forecasting, ensemble forecasting, and integration approaches were all presented to demonstrate the range of forecasting capabilities \cite{hong_2014_gefcom2012}. Gradient boosting machines, semi-parametric models, multiple linear regression, neural networks, random forests, and additive models \cite{hong_2014_gefcom2012} were considered to be all winning techniques for the load forecasting track. In addition, weather station selection and recency effect \cite{wang_2016_recency} were also shown to effectively improve forecasting performances.

Conventional STLF pose some challenges today for independent system operators (ISOs) and utilities because of new operating environments and technologies, including increased penetration of behind the meter distributed energy resources (DERs) \cite{zhehan_2017_fault, yachen_2017_load}, use of new demand side management tools \cite{desong_2015_human, yongli_2014_load}, and the prevalence of microgrids \cite{jiajun_2016_fault, chen_2017_microgrid}. In these situations, traditional point forecasting cannot adequately capture uncertainty, a task that is better accomplished by probabilistic load forecasting (PLF). PLF refers to predicting load in the form of intervals, density functions, or other probabilistic structures instead of a single point output. Compared to the point forecasting, probabilistic forecasting is a better alternative due to the fact that it can reveal more information on the uncertainty. As the point forecast is never perfect, probabilistic forecasting approaches in general are particularly appealing and comprehensive for the power system decision-making process as more and more uncertainties are involved in dynamic environments \cite{hong_2016_review}.

Many researchers have contributed their knowledge and expertise to the probabilistic load forecasting field. In \cite{hong_2016_review}, Hong and Fan summarize existing approaches and provide an extensive overview of PLF. Xie \emph{et al.} \cite{xie_2018_variable} discuss the variable selection process of probabilistic forecasting and propose a holistic method based on probabilistic error metric. Liu \emph{et al.} \cite{liu_2017_sister} propose a quantile regression averaging method to employ a set of sister point forecasting results. Wang \emph{et al.} \cite{yi_2018_residual} conduct the residual quantile fitting after obtaining the point forecast results. Wang \emph{et al.} again in another paper \cite{yi_2018_combine} build an ensemble PLF models by combining with a CQRA method. The same group \cite{yi_2019_lstm} also presents a work to apply huber norm to differentiate the original pinball loss with long-short term memory deep neural network (LSTM). Taieb \emph{et al.} \cite{taieb_2017_hierarchical} propose a hierarchical probabilistic forecasting framework to predict the smart meter demand data and maintain the load hierarchy. 

Motivated by these previous successes, the Global Energy Forecasting Competition 2014 (GEFCOM2014) \cite{hong_2016_gefcom2014} was held to encourage new PLF methods that can be applied to load, price, wind, and solar. This competition led to several high quality papers and techniques, including residual simulation \cite{xie_2016_residual}, additive models with robust aggregation \cite{gallard_2016_additive}, and lasso estimation \cite{ziel_2016_lasso}. 

This paper makes the following contributions: 
\begin{itemize}
    \item A two-stage probabilistic load forecasting approach is proposed to integrate the point forecast as key inputs for the probabilistic forecasting. In the first stage, point forecasting is conducted to provide the load forecast with additional features to enable second stage forecasting and to be able to select features based on feature importance. Then, the second stage combines the point forecast and selected features to efficiently generate the probabilistic forecast with desired quantile levels. 
    
    \item A detailed case study based on ISO New England load data is used to demonstrate the effectiveness of the proposed method in hour-ahead load forecasting. When compared with benchmark cases, the proposed two-stage approach achieves lower forecast errors and narrower prediction intervals.
\end{itemize}

The rest of the paper is organized as follows. Section II presents the detailed method to conduct probabilistic load forecasting. Section III provides the numerical results based on the ISO New England demand to demonstrate the effectiveness of the proposed approach. Section IV concludes the paper.

\section{Method}

This section first introduces the quantile method. Next, the two-stage PLF framework is proposed, including the point forecasting model in stage-1 and probabilistic forecasting model in stage-2. The feature selection and evaluation metrics are presented at the end of this section.

\subsection{Quantile Method}

Traditional load forecasting minimizes the $\ell_2$-norm to provide the conditional mean $\hat{y}_{t}$ of target $y_{t}$ as shown in equation \eqref{eq:l2}, and only a single output is given. 

\begin{equation} \label{eq:l2}
    \mathcal{L} (\hat{y}_{t}, y_{t}) = \left\lVert \hat{y}_{t} - y_{t} \right\rVert_{2}
\end{equation}

Probabilistic forecasting, on the other hand, aims at estimating the probability distribution to fully reveal the future uncertainties. One of the most widely acknowledged probabilistic forecasting approaches is to compute a group of quantiles to discretize the density function for the targeted time interval. Quantile function represents the inverse of cumulative density function (CDF). Assuming $Y$ is a real-valued random variable, the CDF $F$ and the corresponding $q$-quantile are given in equations \eqref{eq:cdf} and \eqref{eq:quantile}.

\begin{equation} \label{eq:cdf}
    F_Y(y) = P(Y \leq y)
\end{equation}
\begin{equation} \label{eq:quantile}
    Q_Y(q) = F_Y^{-1}(q)= \inf \{ y | F_Y(y) \geq q\}
\end{equation}

In probabilistic forecasting, pinball loss function is commonly adopted to evaluate the estimation performances as shown in \eqref{eq:pinball_loss}:
\begin{equation} \label{eq:pinball_loss}
\mathcal{L}_{q} (\hat{y}_{t,q}, y_{t}) = 
\left\{
             \begin{array}{lr}
             (1 - q) (\hat{y}_{t,q} - y_{t}), \quad \hat{y}_{t,q} \geq y_{t} \\
             q (y_{t} - \hat{y}_{t,q}), \quad \hat{y}_{t,q} < y_{t}\\
             \end{array}
\right.
\end{equation}
where $\hat{y}_{t,q}$ is the estimated $q$-quantile output.

For the quantile regression problem, $\hat{y}_{t,q}$ is represented with a linear form as in equation \eqref{eq:quantile_reg}, where $X_t$ and $\beta_{q}$ are the feature vector and the estimated parameters for the quantile level-$q$, respectively.
\begin{equation} \label{eq:quantile_reg}
    \hat{y}_{t,q} = X_t \beta_{q}
\end{equation}

Similar to the linear quantile regression form as in \eqref{eq:quantile_reg}, $\hat{y}_{t,q}$ can be estimated with other forms as well to minimize the pinball loss. For example, quantile regression neural network (QRNN), quantile gradient boosting regression (QGBR) and quantile regression forests (QRF) are all applicable in this task.

\subsection{Two-stage PLF Approach}

\begin{figure}[htb]
    \includegraphics[width=\columnwidth]{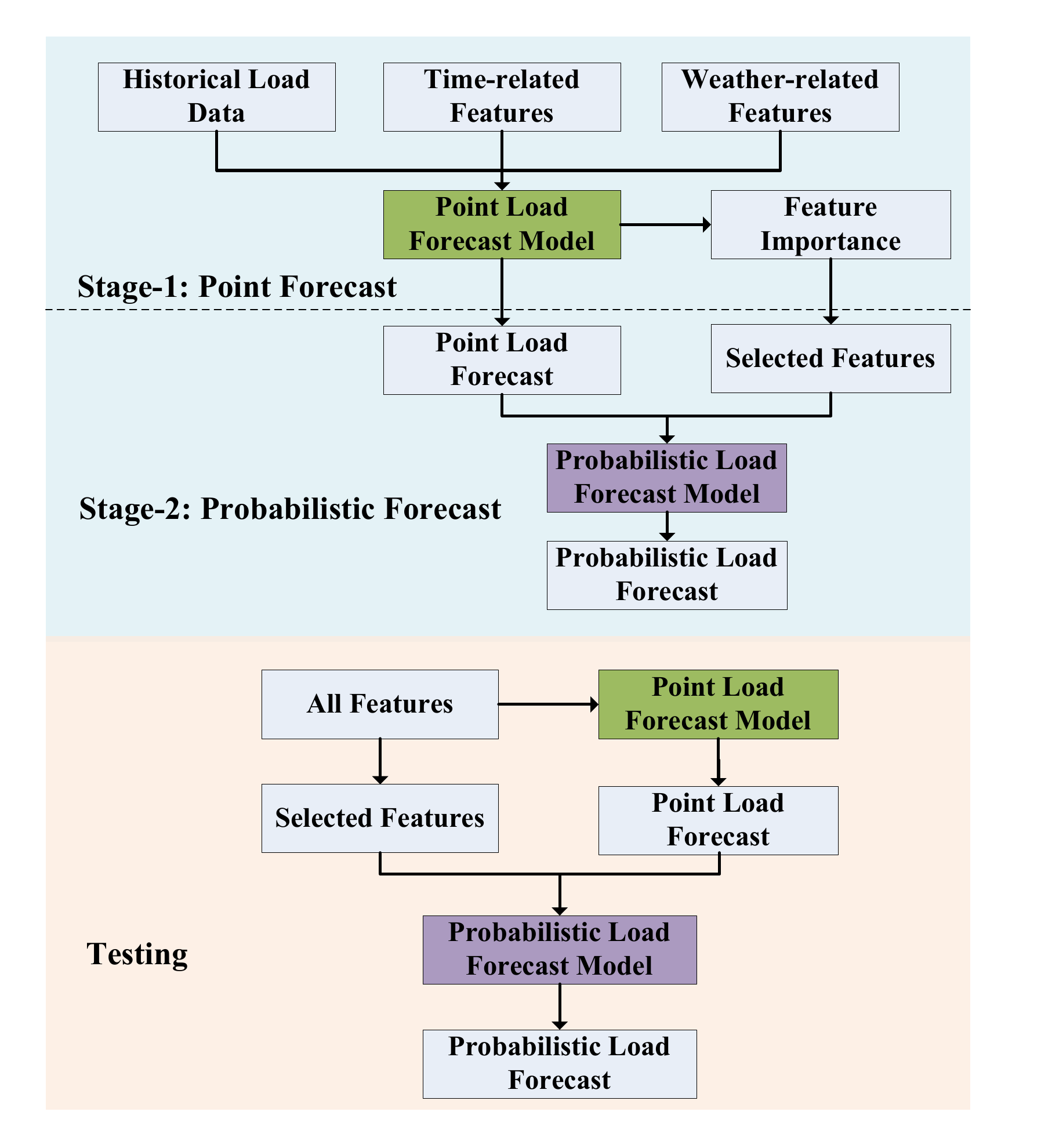}
    \centering
    \caption{Framework of the proposed two-stage method}
    \label{fig:framework}
\end{figure}

The input data for the proposed two-stage approach include historical demand data, time-related features, and weather-related features. Time-related features are generated with one-hot encoding to represent the binary indicators of the month of year, day of week, and hour of day. Weekdays and weekends are thus not differentiated. Regarding the weather related features, dry-bulb temperatures and dew point temperatures are collected from the weather stations. Dry-bulb temperature is actually the ambient air temperature, and relative humidity is measured by the degree of closeness between the dry bulb temperature and the dew point temperature. The historic air temperature and relative humidity are considered in the model training to account for the recency effect as suggested in \cite{wang_2016_recency}. Equation \eqref{eq:humid} computes this relative humidity:
\begin{equation} \label{eq:humid}
    H_{t} = 100 - (T^{db}_{t} - T^{dp}_{t})
\end{equation}
where $H_{t}$, $T^{db}_{t}$, $T^{dp}$ represent the relative humidity, dry bulb temperature, and dew point temperature at time $t$, respectively. Higher values suggest a higher air humidity. 

Fig.~\ref{fig:framework} presents the flowchart of the proposed two-stage probabilistic load forecasting approach. The training process is split into two stages: stage-1 and stage-2. In stage-1, the input features are first used to train a point forecasting model, which provides the feature importance and point forecast outputs to stage-2. The features are ranked according to the contributions to the forecasting results, which are the outputs from tree-based regression methods, such as gradient boosting regression (GBR). These selected high-impact features aim to reduce the stage-2 computing time by extracting necessary information to ensure solution quality. In stage-2, these selected features as well as the produced point forecast are then fed into the probabilistic forecasting engine to train the model. 

In the testing process, test data is first fitted into the trained stage-1 point forecasting model; then the output and the selected features from stage-1 are used by the trained stage-2 probabilistic forecasting model to generate final quantile predictions.

Various machine learning \cite{hong_2014_gefcom2012, hong_2016_gefcom2014} methods can be incorporated into stage-1 and stage-2 model training. For instance, random forests, gradient boosting regression (GBR) and deep neural networks (DNN) can be applied for the point forecasting model; QRF, QGBR, and QRNN are possible options for the probabilistic load forecasting model. Other machine learning techniques can also be applied as well. In this paper, GBR is selected for stage-1, and QRNN is selected for stage-2. For benchmark settings, a direct QGBR model and direct QRNN are trained over the whole training stage-1 and stage-2 to generate probabilistic load forecasting for testing.

\subsection{Feature Selection}

In the forecasting model, feature selection is a critical step to determine the machine learning model inputs. The goal is to identify the important feature candidates and to explore the best feature combinations that are adequate for revealing insightful knowledge with the least amount of information redundancy. Rather than directly feeding all features into the model, this feature selection step is also helpful to improve the model computational efficiency with reduced input dimensions. 

In the proposed two-stage PLF framework, all features, including historical load data, time and weather related features, are first used in the stage-1 point load forecasting model. GBR is applied in this stage for feature selection as it can produce the relative feature importance for all input features. With such a procedure, the most important features are identified through a list of features ranked by their relative importance rate. Then, the cumulative importance cut point is defined to determine which feature combination to adopt in stage-2. In the meantime, the point load forecast given by stage-1 is also used as an additional input feature for the second stage. Via point forecast integration, a set of new feature combinations is constructed that significantly reduces the input dimension for the second stage model while retaining the most information.

 Other feature selection methods, including lasso regression, ridge regression or forward selection, can also be applied. GBR provides a more interpretable way to combine due to its prediction process, which is why it is adopted in the approach presented in this paper.

\subsection{Evaluation Metric}

In point forecasting settings, common evaluation metrics, including root-mean-square error (RMSE), mean-absolute error (MAE), and mean-absolute-percentage error (MAPE), are applied to assess the prediction accuracy. However, these methods are not suitable for evaluating probabilistic forecasting. In this paper, the main metric used is pinball loss as in equation \eqref{eq:pinball_loss}, which provides a comprehensive evaluation. All of the interested quantile levels $0.05, 0.10, 0.15, \cdots, 0.95$ would be estimated. The average pinball loss across all the observations and quantile levels is then calculated as the final evaluation result.

Another probabilistic evaluation metric adopted in this paper is Winkler Score, which is based on both the coverage and the width of prediction intervals. Defined as follow:
\begin{equation} \label{eq:ws}
WS_t = 
\left\{
             \begin{array}{lr}
             \delta_t, \quad & U_t \geq y_t \geq L_t \\
             \delta_t + 2(L_t - y_t)/ \alpha, \quad & L_t > y_t \\
             \delta_t + 2(y_t - U_t)/ \alpha, \quad & U_t < y_t \\
             \end{array}
\right.
\end{equation}
where $\delta_t$, $L_t$, $U_t$, $\alpha$ are the width, the lower bound, the upper bound, and the confidence levels of prediction intervals, respectively.

In addition, prediction interval coverage probability (PICP) is also computed to assess quantile predictions:
\begin{equation} \label{eq:picp_c}
c_{i}=
\left\{
             \begin{array}{lr}
             1, \quad & y_{i} \in I_{t}^{\alpha} (x_{i}) \\
             0, \quad & y_{i} \notin I_{t}^{\alpha} (x_{i})\\
             \end{array}
\right.
\end{equation}
\begin{equation} \label{eq:picp}
PICP = \frac{1}{N} \sum_{i=1}^{N} c_{i}
\end{equation}
where $c_{i}$ indicates whether the $i$-th actual load value $y_{i}$ is included in the interested $\alpha$-level prediction interval, and $N$ is the number of testing samples. Therefore, a PICP larger than $\alpha$ implies a reliable forecasting result.

\section{Case Study}

\subsection{Simulation Setup}

The proposed forecasting approach has been tested with the load data from ISO New England public dataset \cite{isone_load}, including eight sub-load zones (CT, ME, NH, RI, VT, SEMA, WCMA and NEMA) and the total system load. Hourly demand and weather information from 2013-1-1 to 2017-12-31 are collected. Particularly, the first three-year data is used for training and validation for stage-1, and the fourth year is used for stage-2. Testing stage is conducted with the last year data. 
\begin{figure}[htb]
    \centering
    \includegraphics[width=0.85\columnwidth]{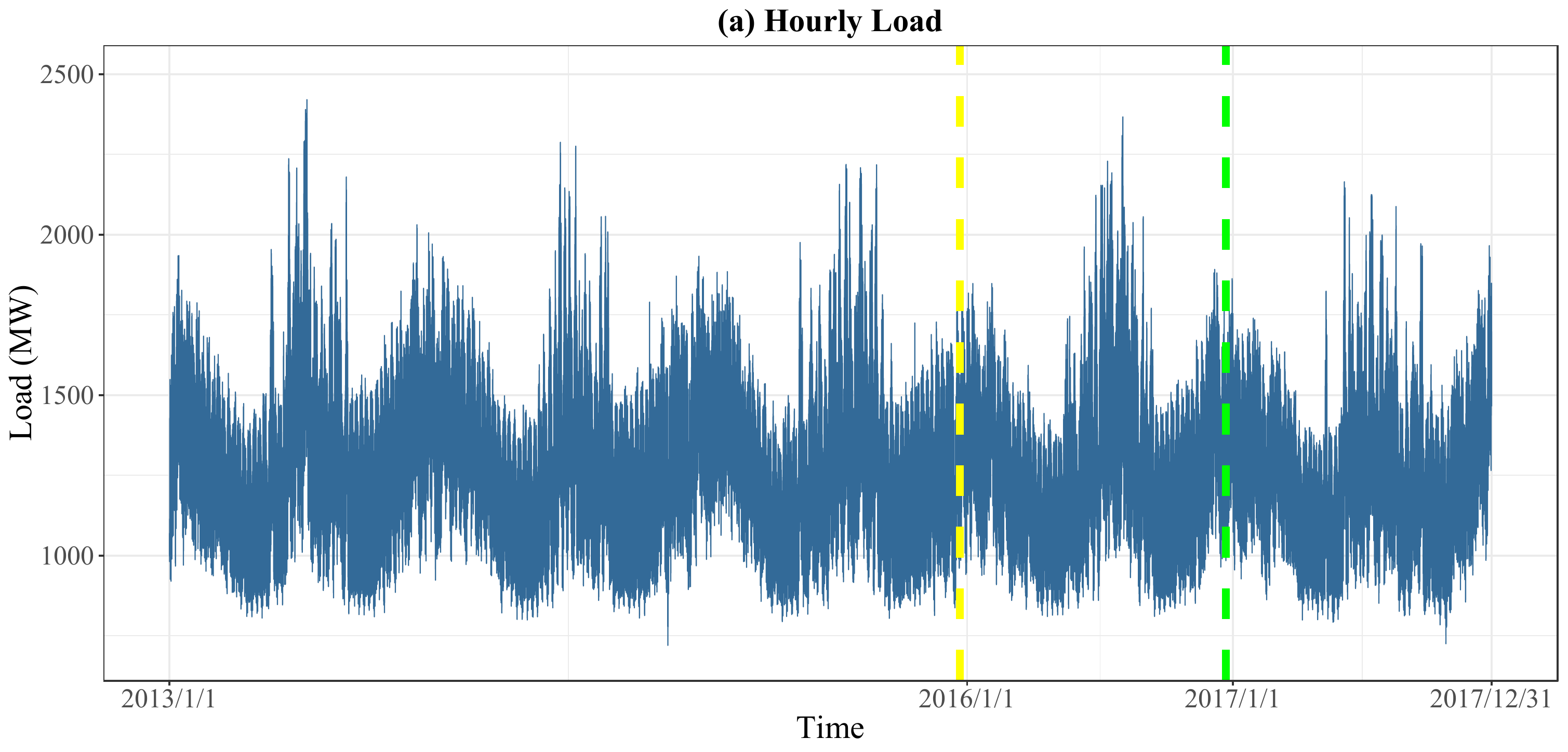}
    \includegraphics[width=0.85\columnwidth]{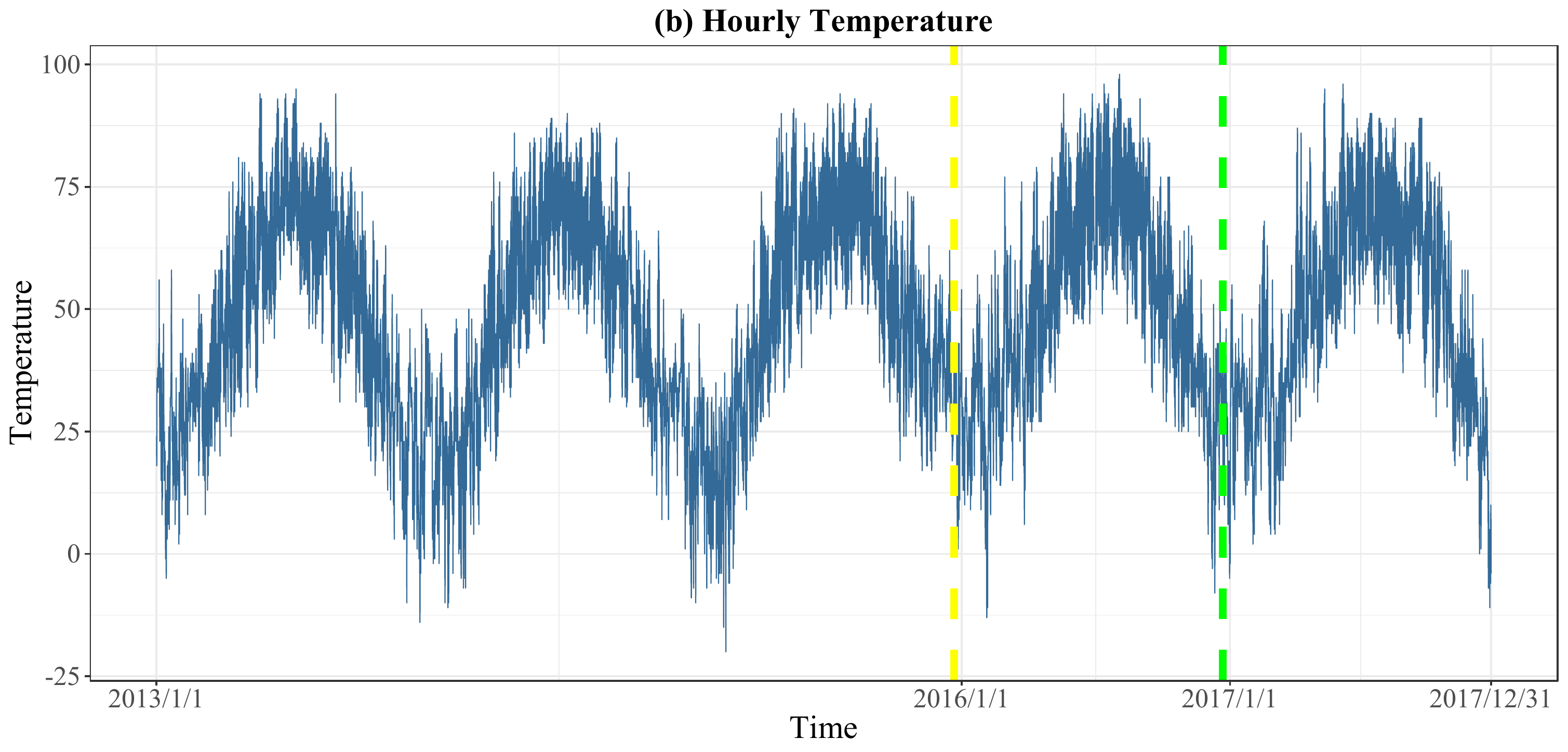}
    \includegraphics[width=0.85\columnwidth]{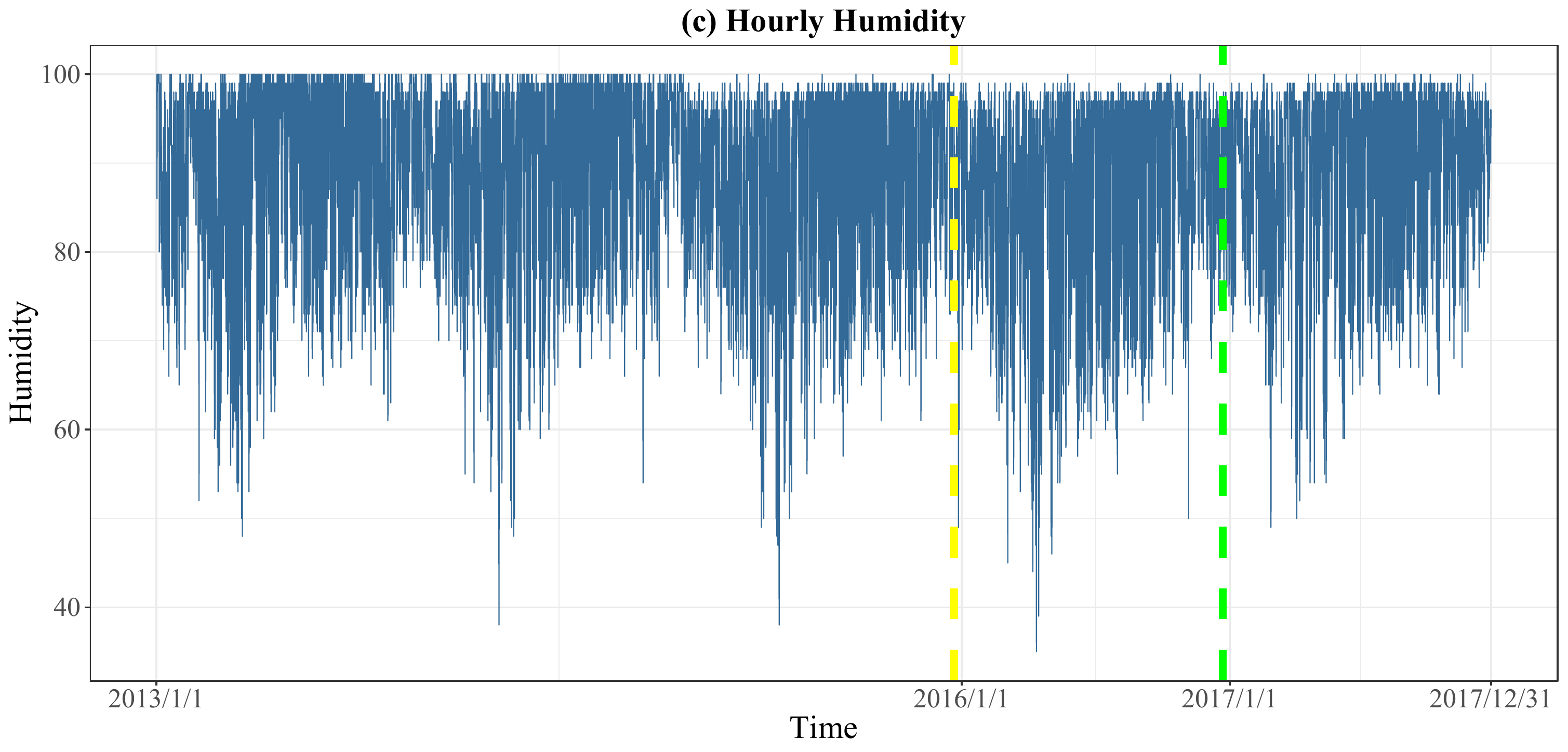}
    \caption{New Hampshire historical data: 2013 -- 2017}
    \label{fig:demand}
\end{figure}

For illustration purpose, Fig.~\ref{fig:demand} presents the hourly demand, temperature (dry bulb) and relative humidity in the New Hampshire (NH) area. The yellow and the green dot line separate stage-1, stage-2, and the testing stage. Seasonal and periodic patterns are easily identified for the load and temperature.

\begin{figure}[htb]
    \includegraphics[width=0.85\columnwidth]{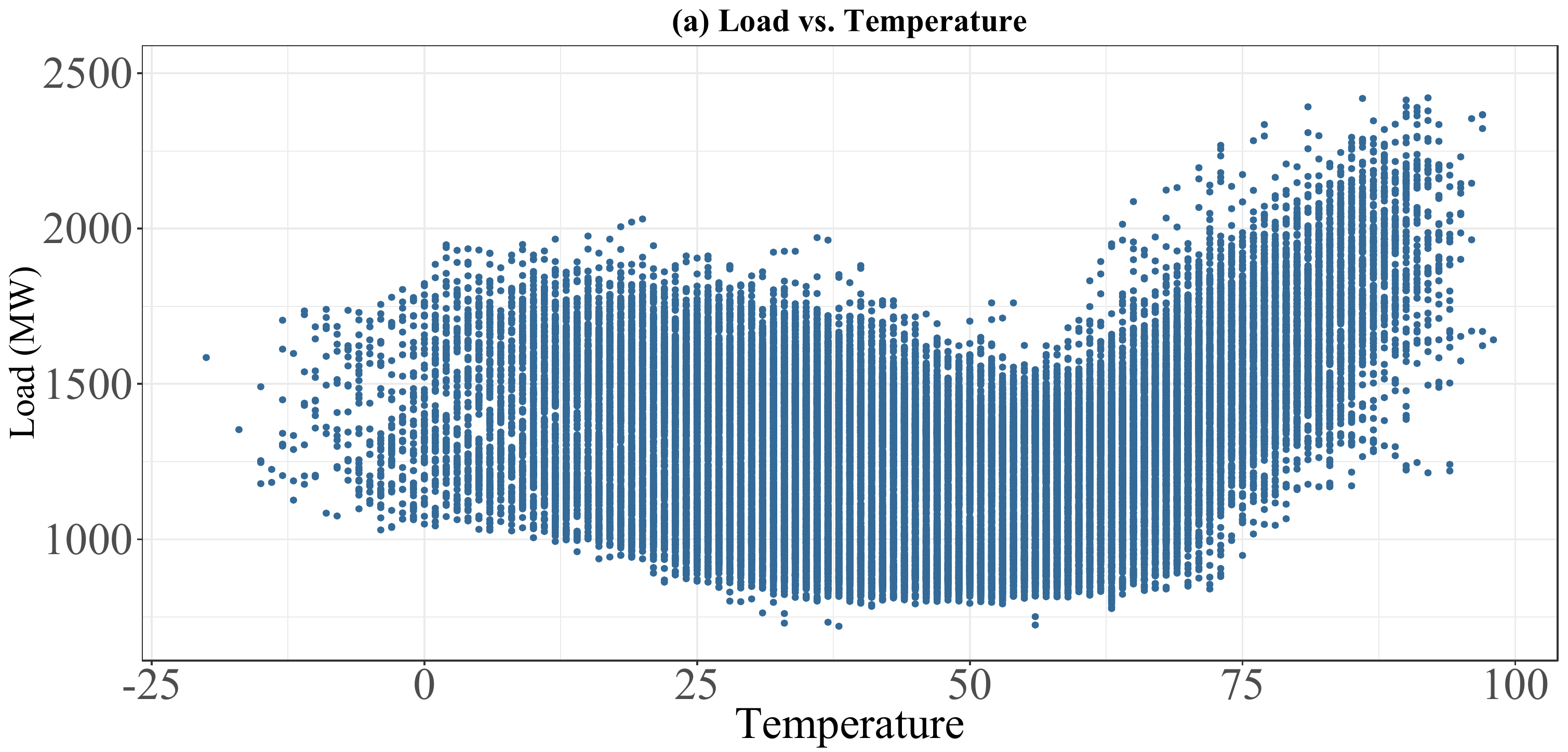}
    \includegraphics[width=0.85\columnwidth]{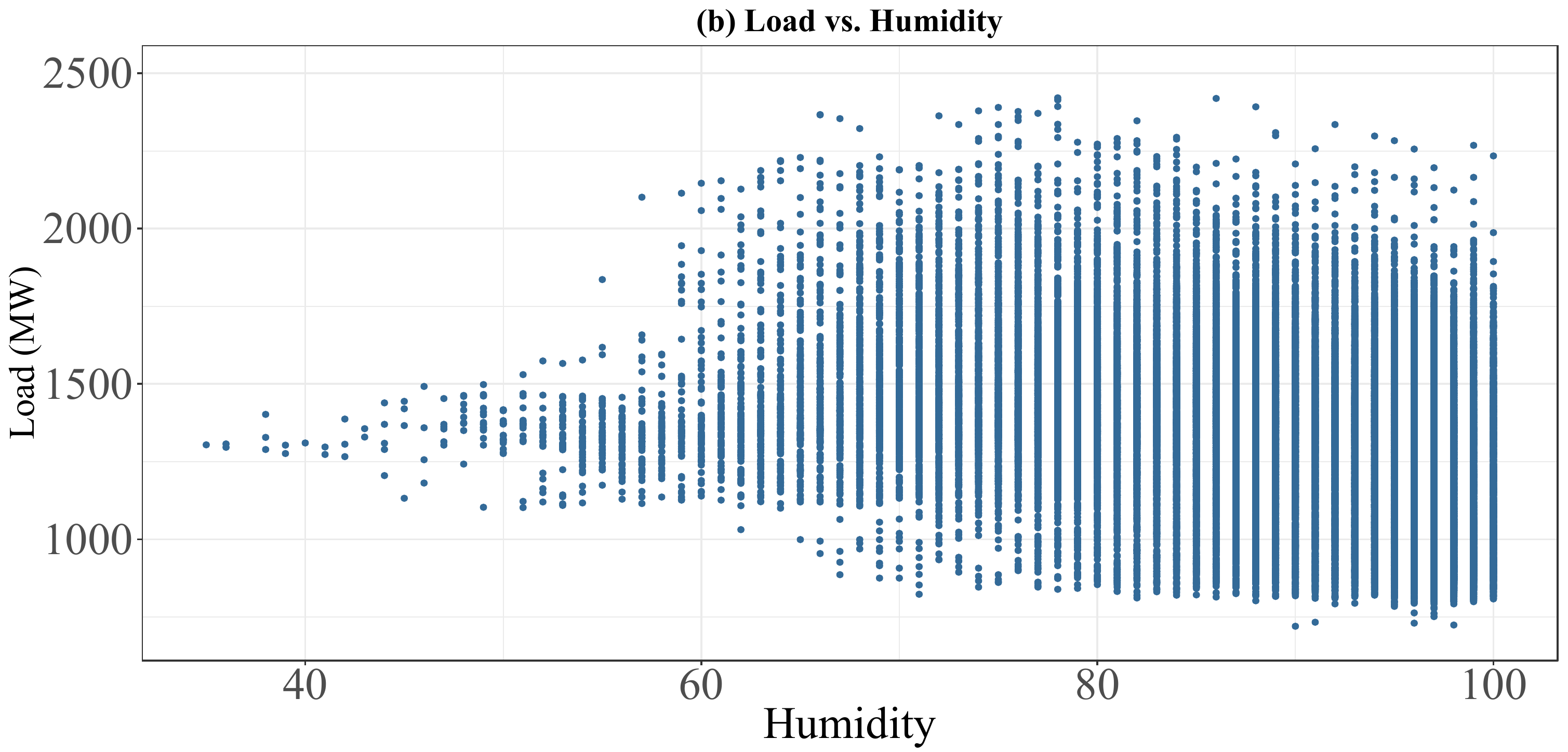}
    \centering
    \caption{New Hampshire Weather Impacts: 2013 -- 2017}
    \label{fig:demand_temp}
\end{figure}

Fig.~\ref{fig:demand_temp} shows the scatter plots for (a) demand versus temperature and (b) demand versus relative humidity in the New Hampshire area. In Fig.~\ref{fig:demand_temp}(a), the 'V' shape indicates a strong correlation between the demand and the temperature. Therefore, it is important to include such effect in the model. On the other hand, no clear relationship is identified between the demand and relative humidity, as shown in Fig.~\ref{fig:demand_temp}(b).

In this paper, hour-ahead forecasting is conducted with the proposed two-stage PLF method on all eight load zones and the entire ISO-New England. Quantiles in $5\%, \cdots, 95\%$ are predicted for the probabilistic forecasting. In addition, medians (50\% quantile) are also provided for the final point measurement. 

All simulations were carried out in R on an Intel Core 4.00 GHz processor with 12.0 GB of RAM.

\subsection{Numerical Results}

\begin{figure}[htb]
    \includegraphics[width=0.85\columnwidth]{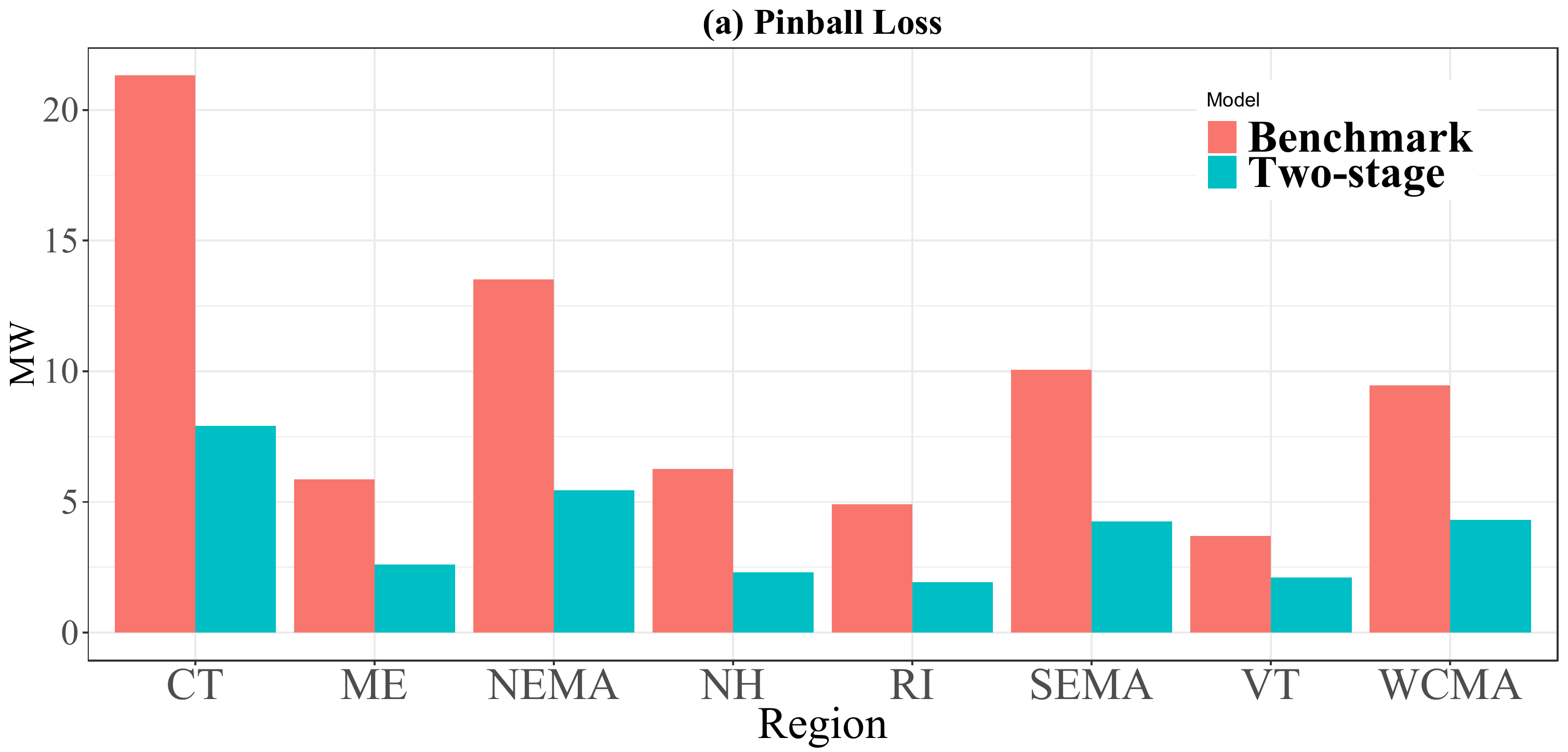}
    \includegraphics[width=0.85\columnwidth]{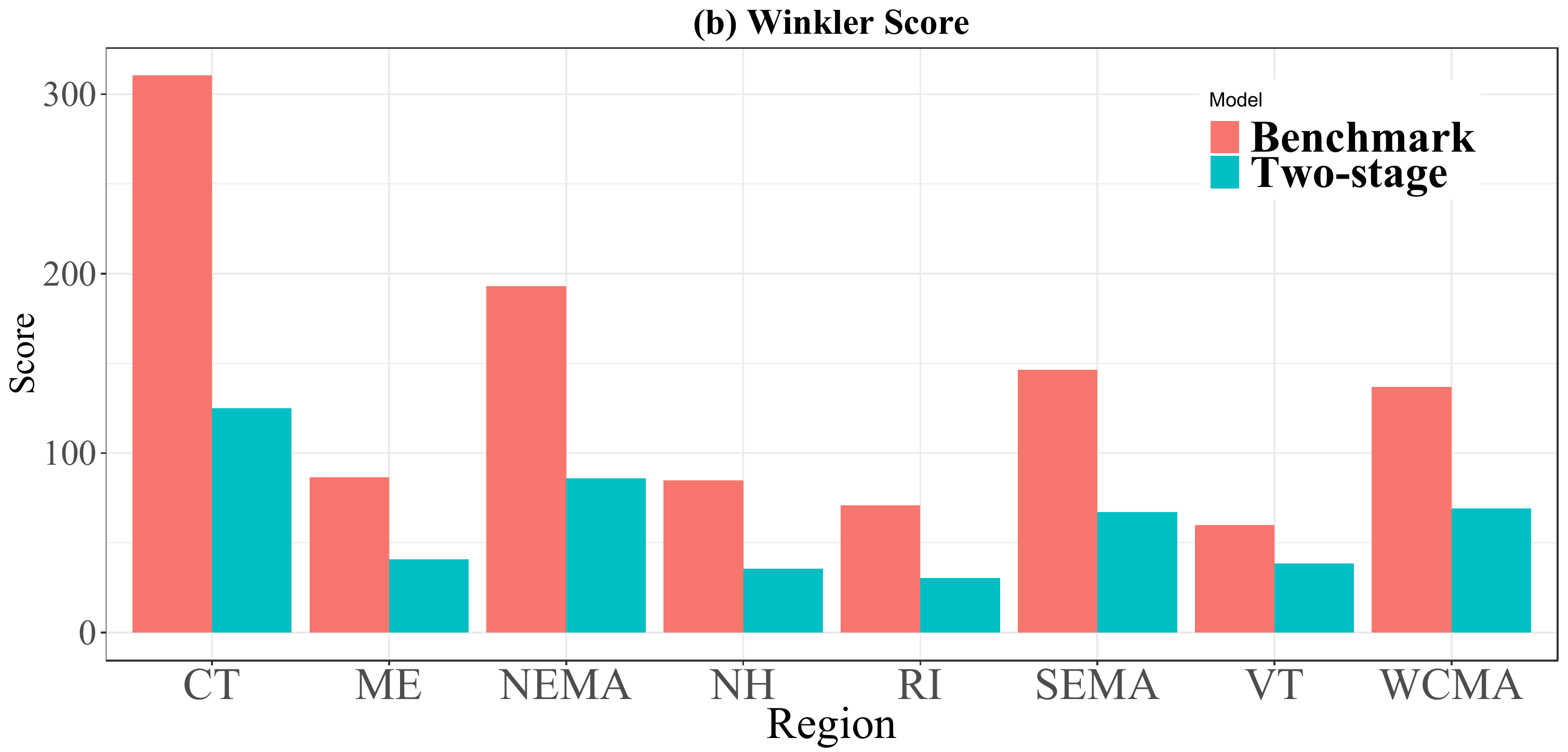}
    \centering
    \caption{Pinball Loss and Winkler Score over ISO New England}
    \label{fig:pin}
\end{figure}

Fig.~\ref{fig:pin} presents the pinball loss and Winkler score over eight load zones. The proposed two-stage method GBR+QRNN is compared with the benchmark method of direct QGBR. The proposed method outperforms the benchmark in all areas to improve the forecasting accuracy. The New Hampshire load zone is selected for further illustrations.

Table.~\ref{tab:models} presents the probabilistic load forecasting evaluation by MAE, RMSE, prediction interval width between 5\% and 95\% levels, averaged Pinball loss, averaged Winkler score, and PICP for direct models and two-stage models. For two-stage methods in stage-2, different QRNN structures are also explored to further optimize its structure. Also, improvement rates are given based on pinball loss. From Table~\ref{tab:models} and Fig.~\ref{fig:pin}, the proposed two-stage PLF framework achieves a huge improvement in quantile prediction accuracy. The reason is that integrating the point forecast in the feature greatly improves the probabilistic forecasting model capabilities by explicitly capturing the future load behaviors. In addition, it implicitly incorporates the raw features in stage-1 for point forecasting, which requires less features in the stage-2 for probabilistic forecasting. It clearly shows the simple yet effective effect with such point forecast integration.

\begin{table*}[htb]
\caption{Model Comparison for Testing Evaluation Metrics}
\label{tab:models}
\centering
\begin{tabular}{c|c|c|c|c|c|c|c}
\hline
\hline
\textbf{Model}  & \textbf{MAE} & \textbf{RMSE} & \textbf{PI Width} & \textbf{Pinball Loss} & \textbf{Winkler Score} & \textbf{PICP} & \textbf{Improvement Rate} \\ \hline \hline
Direct QGBR     & 8.66         & 23.57         & 82.25             & 6.25                  & 84.86                  & 0.92          & 0                         \\
Direct QRNN     & 6.55         & 18.40         & 48.97             & 4.73                  & 69.05                  & 0.84          & 24\%                      \\
GBR+QGBR        & 4.81         & 9.48          & 36.83             & 3.62                  & 58.77                  & 0.77          & 42\%                      \\
GBR+QRNN(5)     & 3.44         & 10.10         & 28.24             & 2.52                  & 37.70                  & 0.84          & 59\%                      \\
GBR+QRNN(10)    & 3.17         & 9.29          & 23.97             & 2.31                  & 35.60                  & 0.82          & 63\%                      \\
GBR+QRNN(10,5)  & 3.02         & 8.87          & 24.45             & 2.33                  & 36.83                  & 0.83          & 62\%                      \\
GBR+QRNN(20,10) & 3.11         & 9.09          & 21.70             & 2.36                  & 39.15                  & 0.76          & 62\%                      \\
GBR+QRNN(30,15) & 3.31         & 9.48          & 19.29             & 2.48                  & 43.22                  & 0.71          & 60\%                      \\ \hline \hline
\end{tabular}
\end{table*}

\begin{figure}[htb]
    \includegraphics[width=0.85\columnwidth]{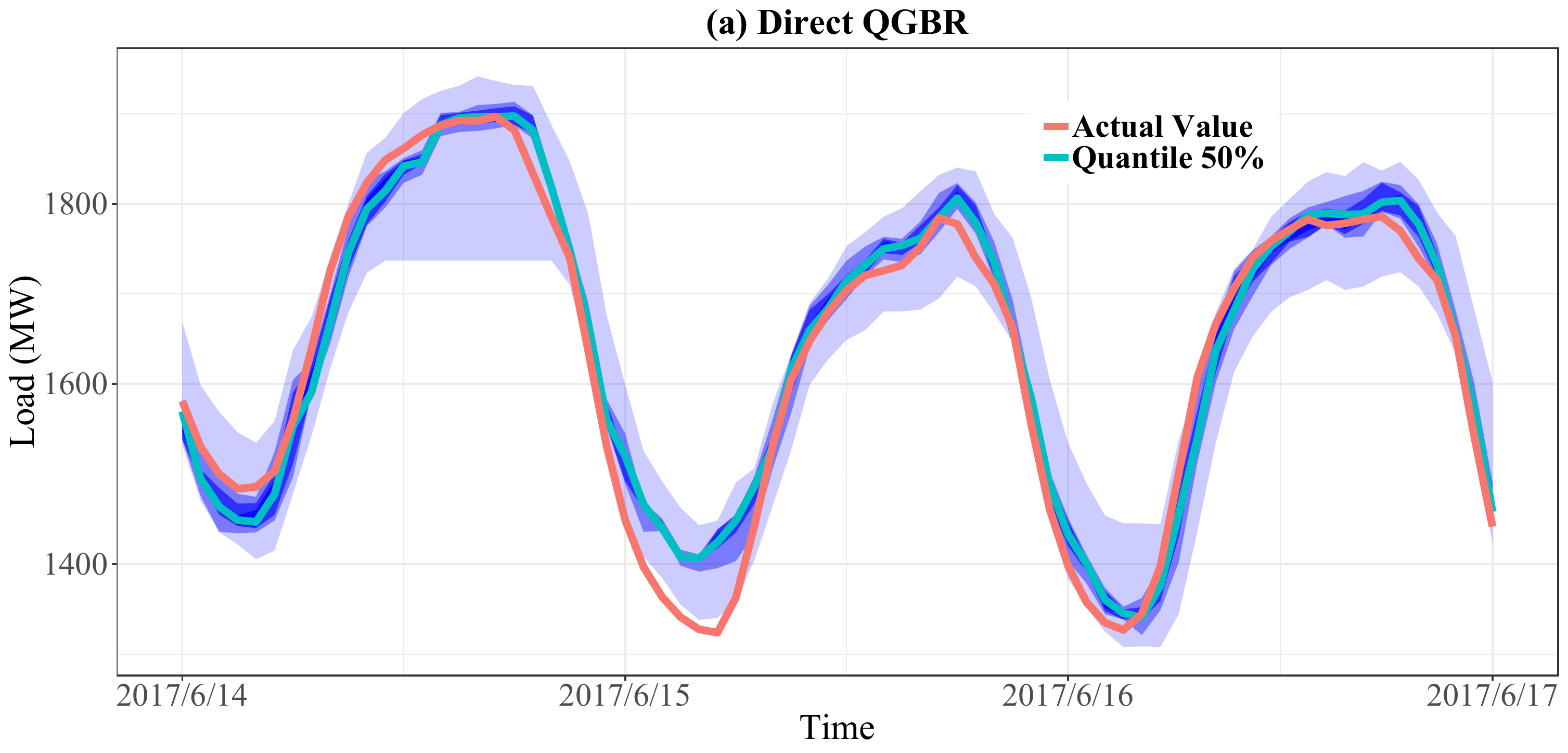}
    \includegraphics[width=0.85\columnwidth]{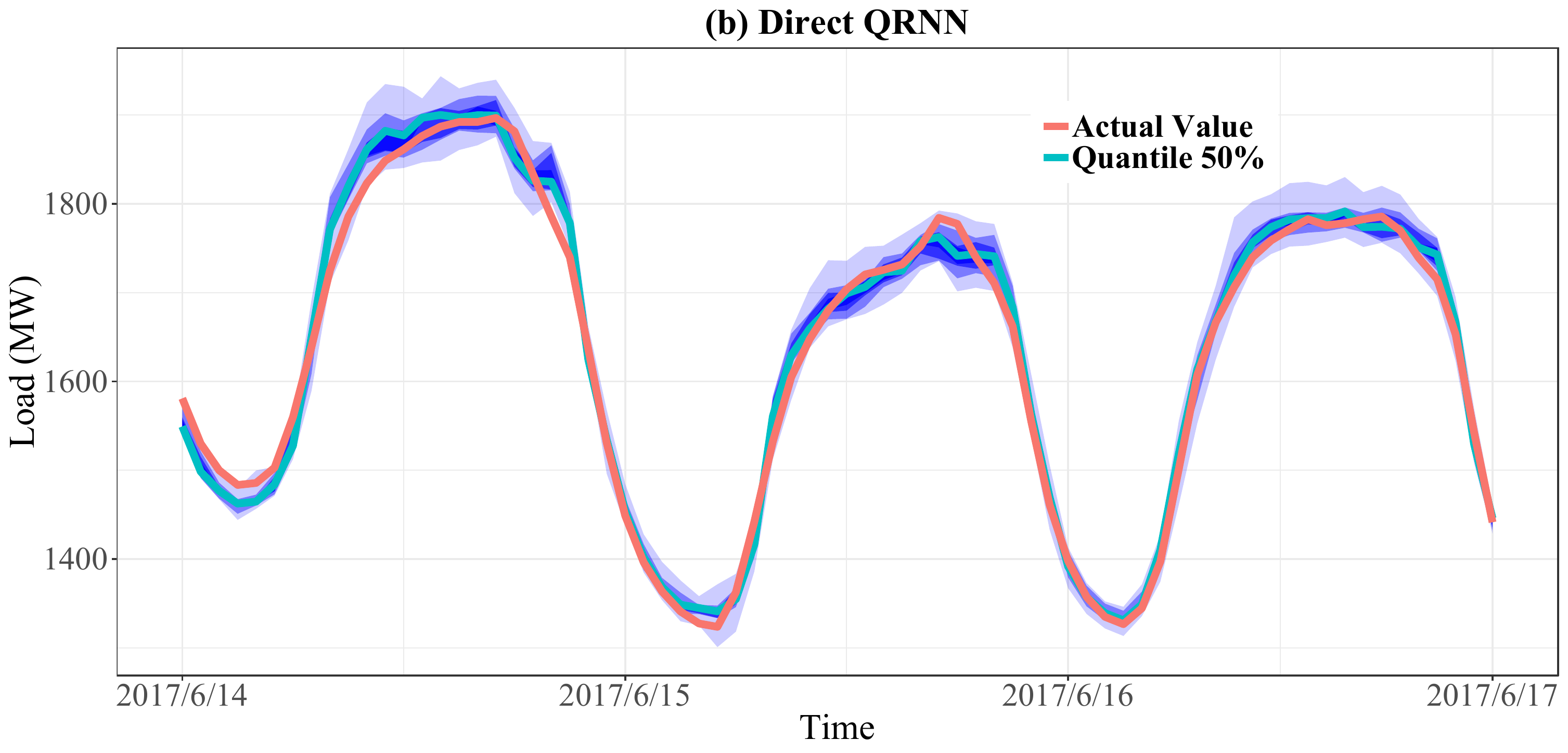}
    \includegraphics[width=0.85\columnwidth]{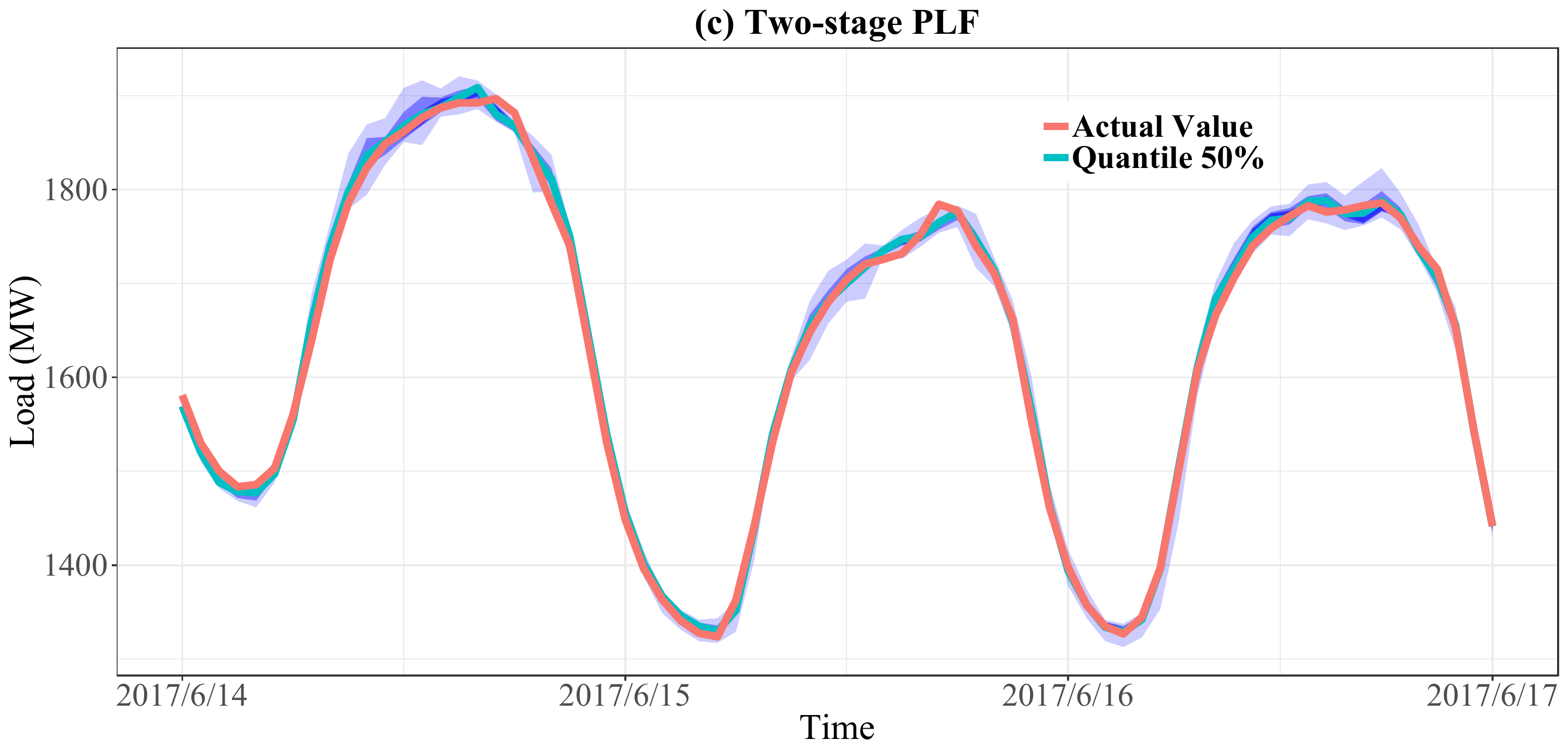}
    \centering
    \caption{Direct QGBR Model for 72-hour Real Time Demand Forecasting from 2017-06-14 00:00 to 2017-06-17 00:00}
    \label{fig:DQGBRday}
\end{figure}

Fig.~\ref{fig:DQGBRday} shows a 72-hour probabilistic load forecast result in the New Hampshire area from 2017-06-14 00:00 to 2017-06-17 00:00 for (a) Direct QGBR, (b) Direct QRNN and (c) two-stage GBR+QRNN. Compared with the benchmark methods, a more narrower prediction interval is clearly demonstrated when using the proposed two-stage method. In addition, the actual loads are almost always within the predicted ranges with the two-stage GBR+QGBR model, whereas there is an off-target period for the benchmark model around 2017-06-15 00:00. Since this is an ISO-level hour-ahead forecasting, the uncertainty range should not be large or wide. The proposed two-stage GBR+QGBR model meets such intuition and requirement. In the future work to extend from the hour-ahead to day-ahead forecasting, this prediction intervals are expected to be wider.

Table.~\ref{tab:feature} shows the selected features from the stage-1 GBR model for ISO New England total load, ranked by their relative importance. The top-ranked features serve as input features for the stage-2 probabilistic forecasting model. As hour-ahead forecasting is the objective, it is not surprising to see the historical load for the past 1-day and past 7-days act as more important factors for the prediction.

\begin{table}[htb]
\caption{Important Features Selected in Stage-1}
\label{tab:feature}
\centering
\begin{tabular}{c|c|c}
\hline
\hline
\textbf{Feature}    & \textbf{Importance} & \textbf{Cumulative Rate} \\ \hline \hline
Demand(t-1)         & 81.54               & 81.54\%             \\
Demand(t-23)        & 6.67                & 88.21\%             \\
Demand(t-167)       & 5.70                & 93.91\%             \\
Demand(t-24)        & 1.45                & 95.36\%             \\
Demand(t-168)       & 1.40                & 96.76\%             \\
Demand(t-22)        & 0.60                & 97.36\%             \\
Temperature(t)      & 0.47                & 97.83\%             \\
Demand(t-143)       & 0.27                & 98.10\%             \\
Demand(t-21)        & 0.14                & 98.24\%             \\
Demand(t-18)        & 0.14                & 98.38\%             \\
Demand(t-26)        & 0.12                & 98.50\%             \\
Demand(t-27)        & 0.11                & 98.61\%             \\ 
Friday              & 0.11                & 98.72\%             \\ \hline \hline
\end{tabular}
\end{table}

\begin{figure}[htb]
    \includegraphics[width=0.8\columnwidth]{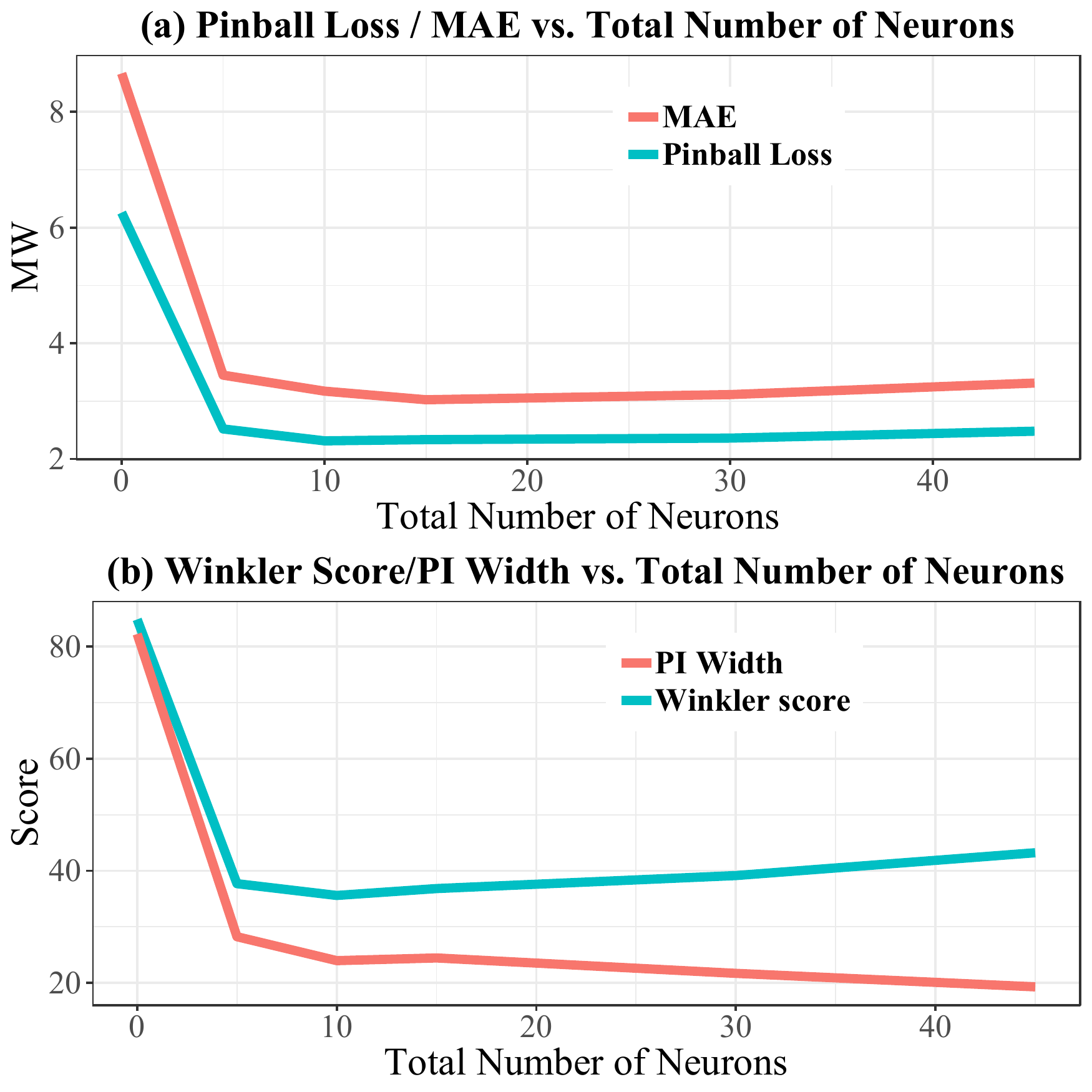}
    \includegraphics[width=0.8\columnwidth]{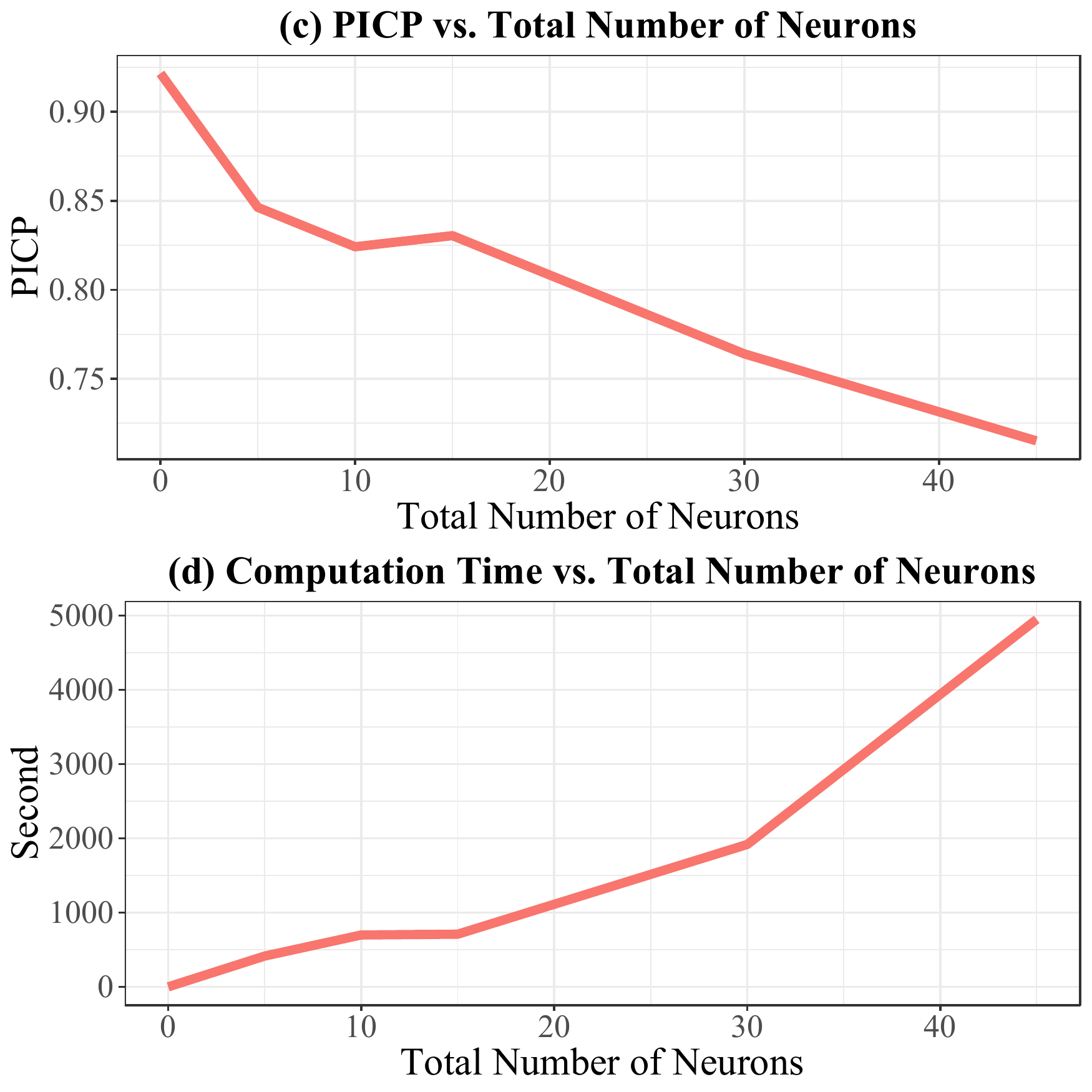}
    \centering
    \caption{Evaluation Metrics for NN structure}
    \label{fig:nn}
\end{figure}

To further investigate the proper neural network structures in the stage-2 probabilistic forecasting, extensive simulations are carried out to explore how they affect forecasting performances as shown in Fig.~\ref{fig:nn}. From these figures, the trends in all Pinball loss, Winkler-Score, and MAE imply that the probabilistic forecasting accuracy is improved to a certain point as the NN structure becomes more complex. The accuracy improvement is then merely marginal after the structure becomes too complex. This suggests a properly optimized structure is important for stage-2 QRNN to avoid overfitting. In this case, the lowest pinball loss and Winkler score are obtained simultaneously at a single-layer NN in 10 neurons while the lowest MAE and RMSE are obtained at a two-layer NN in $[10, 5]$ neurons. Moreover, the decreasing trend of prediction interval width indicates the overfitting issues with complex NN, especially for this highly predictable hour-ahead load data. Another factor is the computational efficiency. Fig.~\ref{fig:nn}(d) shows that the training time significantly increases when more neurons are used, however, with minimal accuracy improvement. Therefore, a single-layer with 10 neuron QRNN model is the appropriate NN structure in this problem to achieve the tradeoff between the accuracy and the computation speed. In addition, this NN structure choice also meets the Occam's razor principle to select simpler models when possible.

\section{Conclusion}

This paper proposes a two-stage probabilistic load forecasting method to integrate point forecast as key forecasting enablers. In the first stage, the point forecasting model provides the point load forecast as the core features in the second stage probabilistic forecasting model. In addition, historical load, time and weather related features are selected by their relative importance rate for the second stage. This predictive framework significantly improves the forecasting accuracy compared with direct quantile forecast. Numerical results from ISO New England load demonstrate the effectiveness of the proposed method on hour-ahead probabilistic load forecasting. Moreover, a relative optimized NN structure of the second stage QRNN model achieves both the  forecasting accuracy as well as the computation efficiency. 

\bibliographystyle{IEEEtran}
\input{output.bbl}

\end{document}